\documentclass[12pt]{article}
\pdfoutput=1
\usepackage{jcappub}
\usepackage{amsmath}
\usepackage{amssymb}
\usepackage{physics}
\usepackage{xcolor}
\usepackage{slashed}
\usepackage{float}
\usepackage{notoccite}
\usepackage{enumerate}
\usepackage{tabularx}
\usepackage{array}
\usepackage{empheq}

\usepackage{graphicx}   
\usepackage{subcaption} 
\usepackage{float}    
\usepackage{placeins}

\def\be{\begin{equation}}
\def\ee{\end{equation}}
\def\bea{\begin{eqnarray}}
\def\eea{\end{eqnarray}}
\def\bes{\begin{subequations}}
\def\ees{\end{subequations}}

\title{Dataset-learning duality and emergent criticality}

\author[1]{Ekaterina Kukleva}
\emailAdd{ekaterina.kukleva.99@gmail.com}\author[1,2]{and Vitaly Vanchurin} 
\emailAdd{vitaly.vanchurin@gmail.com}

\affiliation[1]{Artificial Neural Computing, Weston, Florida, 33332, USA}
\affiliation[2]{Duluth Institute for Advanced Study, Duluth, Minnesota, 55804, USA}

\begin{document}

\abstract{In artificial neural networks, the activation dynamics of non-trainable variables is strongly coupled to the learning dynamics of trainable variables. During the activation pass, the boundary neurons (e.g., input neurons) are mapped to the bulk neurons (e.g., hidden neurons), and during the learning pass, both bulk and boundary neurons are mapped to changes in trainable variables (e.g., weights and biases). For example, in feed-forward neural networks, forward propagation is the activation pass and backward propagation is the learning pass. We show that a composition of the two maps establishes a duality map between a subspace of non-trainable boundary variables (e.g., dataset) and a tangent subspace of trainable variables (i.e., learning). In general, the dataset-learning duality is a complex non-linear map between high-dimensional spaces. We use duality to study the emergence of criticality, or the power-law distribution of fluctuations of the trainable variables, using a toy model at learning equilibrium. In particular, we show that criticality can emerge in the learning system even from the dataset in a non-critical state, and that the power-law distribution can be modified by changing either the activation function or the loss function.}

\maketitle

\section{Introduction}

Dualities provide a highly useful technique for solving complex problems that have found applications in many branches of science, most notably in physics. For example, well-known dualities include electric-magnetic duality \cite{Griffiths}, wave-particle duality \cite{Feynman}, target-space dualities \cite{Polchinski}, Kramers-Wannier Duality \cite{Kramers1, Kramers2}, etc. More recent, but less-known examples include a quantum-classical duality \cite{Vanchurin1}, dual path integral \cite{Vanchurin2}, and duality pseudo-forest \cite{Vanchurin3}, to name a few. The key idea of all physical dualities is to establish a mapping (i.e., a duality mapping) between two systems (e.g., physical theories), which can then be used to study properties (e.g., obtaining solutions) of one system by analyzing the other system, or vice versa.

Perhaps the most well-studied example of a physical duality is the so-called bulk-boundary or holographic duality \cite{Susskind, tHooft}, such as AdS-CFT \cite{Maldacena, Witten1, Witten2, Gubser}. In AdS-CFT, the mapping is established between the bulk, representing a gravitational theory on the anti-de Sitter (AdS) background, and the boundary, representing a conformal field theory (CFT) without gravity. 

Mathematical dualities focus on more formal, abstract transformations preserving algebraic or geometric structures, but also very useful in physics. For instance, dual vector space duality \cite{Axler}, duality of points and lines in projective geometry \cite{Kagan}, Hom and Tensor Duality \cite{Cartan}, etc.

In this article we shall consider a learning duality, which we shall refer to as the dataset-learning duality. It is often convenient to view the dataset as representing the states of non-trainable variables on the `boundary', and learning as representing the dynamics (or changes of states) of trainable variables in the `bulk'. Thus, the dataset-learning duality may be considered as an example of a bulk-boundary duality in the context of learning theory. As we shall argue, the duality is a complex non-linear mapping between very high-dimensional spaces, but near equilibrium state, the analysis is greatly simplified. This simplification allows us to apply the dataset-learning duality to study the emergence of criticality in learning systems.

Criticality in physical systems refers to the behavior of these systems at or near critical points, where they undergo phase transitions. These critical points are characterized by dramatic changes in physical properties due to the collective behavior of the system's components. Understanding phase transitions and criticality is essential for explaining many physical phenomena \cite{Landau} as well as complex biological phenomena such as biological phase transitions \cite{Romanenko}. Phenomenon of self-organized criticality was investigated using biologically inspired discrete network models, which adapts its topology based on local rules without any central control to achieve a balance between stability and flexibility. In papers \cite{Rohlf, Rybarsch, Landmann, DelPapa, KalleKossio, Cowan} networks reach a critical state, where it exhibits power-law distributions of avalanches sizes and durations, indicative of self-organized criticality (SOC), where the criticality is described by a scale-invariant and power-law distribution of fluctuations. Recent studies have shown that many physical \cite{WNN, Katsnelson, Vanchurin4} and biological \cite{VWKK, VWKK2} systems can be modeled as learning systems, such as artificial neural networks. Therefore, understanding the criticality and phase transitions in artificial neural networks may also shed light on the emergence of criticality in physical and biological systems.

The first step in this direction was made in Ref. \cite{SOC}, where the criticality, or a power-law distribution of fluctuations, was derived analytically using a macroscopic, or thermodynamic, description developed in \cite{TTML}. This results was then confirmed numerically \cite{SOC} for a classification learning task involving handwritten digits \cite{MNIST}. In this paper, we take another step towards developing a theory of emergent criticality by providing a more microscopic description of the phenomena using the dataset-learning duality. We then test our predictions numerically for a toy-model classification task.

The paper is organized as follows. In Sec. \ref{sec:septuple} a theoretical model of neural networks and basic notations are introduced. Sec. \ref{sec:duality} is devoted to developing a statistical description of a local dataset-learning duality. In Sec. \ref{sec:jacobian} the distribution of the variables of the tangent space dual to the boundary (dataset) space is analyzed in the multidimensional case. In Sec. \ref{sec:toy} a toy-model with two trainable and two non-trainable variables is introduced, and in Sec. \ref{sec:criticality} it is solved for power-law fluctuations of the trainable variables. In Sec. \ref{sec:numeric} we present numerical results for some specific power-law distributions obtained for specific compositions of activation and loss functions.  In Sec. \ref{sec:Results} we summarize and discuss the main results of the paper.  

\section{Neural networks}\label{sec:septuple}

Consider an artificial neural network defined as a neural septuple  $({\bf x}, \hat{P}, p_\partial, \hat{w}, {\bf b}, {\bf f}, H)$ \cite{TTML} , where:
\begin{enumerate}
\item ${\bf x} \in \mathbb{R}^N$,  is a vector state of $N$ neurons,  
\item $\hat{P}$,  is a projection to subspace of $N_\partial = \Tr(\hat{P})$ boundary neurons, 
\item $p_\partial({\bf x}_\partial)$, is a probability distribution which describes the training dataset, 
\item $\hat{w} \in  \mathbb{R}^{N^2}$, is a weight matrix which describes connections between neurons, 
\item ${\bf b} \in \mathbb{R}^N $, is a bias vector which describes bias in inputs of individual neurons, 
\item  ${\bf f}(\hat{w} {\bf x} +{\bf b} )$, is an activation map which describes a non-linear part of the dynamics,
\item  ${H}({\bf x}, \hat{w}, {\bf b})$, is a loss function of both trainable and non-trainable variables.
\end{enumerate} 
It is assumed that the bias vector ${\bf b}$  and weight matrix $\hat{w}$ are the only trainable parameters which can be combined into a single trainable vector ${\bf q}$ via transformation tensors $W_{i j}^l$, and $B_i^l$,
\footnote{Einstein summation convention over repeated indices is implied here and throughout the manuscript unless stated otherwise.}
\bea
\label{single_vector}
    w_{i j} &=& W_{i j}^l q_l,\\
    b_i &=& B_{i}^l q_l.\notag
\eea

In standard neural networks, including feedforward \cite{FNN}, convolutional \cite{CNN}, auto-encoder \cite{AE}, transformers \cite{Transformer}, etc., certain trainable variables can be shared, fixed, or set to zero, but all such architectures can be described using appropriate choices of constant tensors  $W_{i j}^l$ and $B_i^l$. Then Eq. \eqref{single_vector} can be viewed as a linear map from $K$-dimensional space of trainable variables to $N^2 + N$-dimensional space of weights and biases, where $K$ can be much smaller than $N^2 + N$.

The neural septuple  $({\bf x}, \hat{P}, p_\partial, \hat{w}, {\bf b}, {\bf f}, H)$ defines the three relevant types of dynamics and their relevant time-scales:
\begin{itemize}
\item Activation dynamics describes changes of bulk neurons ${\bf x}_\slashed{\partial} = (\hat{I} - \hat{P}) {\bf x}$ on the smallest time scale which can be set to one, 
\be
{\bf x}_{\slashed{\partial}}(t+1) = (\hat{I} - \hat{P}) {\bf x}(t+1) = (\hat{I} - \hat{P}) {\bf f}(\hat{w} {\bf x}(t) + {\bf b}),
\ee
where $\hat{I}$ is the identity matrix.

\item Boundary dynamics describes updates of boundary neurons ${\bf x}_\partial = \hat{P} {\bf x}$ on the intermediate time-scales, e.g. once per $L$ unit time steps, e.g. drawn from probability distribution $p_\partial({\bf x}_\partial)$ which describes the dataset.

\item Learning dynamics describes changes of trainable variables ${\bf q}$ on the largest time-scale $M L$ where $M$ is the so-called mini-batch size. For example, for the stochastic gradient descent method, 
\be
\dot{q}_i = {q}_i (t+ML) - {q}_i(t) = -\gamma \delta_{i j} \frac{d \langle  {H} \rangle_M }{d q_j}\label{eq:learning}
\ee
where $\langle  ... \rangle_M$ is the averaging over mini-batch and $\gamma$ is the learning rate. Note that $ML$ is a unit on the scale of ${\bf q}$ changes, in further analytical reasoning, for simplicity, we set $M = 1$. We emphasize that the method \eqref{eq:learning} implies that the space of trainable parameters is flat globally and coordinates $q_i$ are orthonormal. That is, in the case under consideration $g_{i j} = \delta_{i j}$, where $\delta_{i j}$ is the Kronecker delta.
\end{itemize}
For example, in the case of the feedforward neural network the weight matrix $\hat{w}$ must be nilpotent and if its degree is $L$, i.e. $\hat{w}^{L}=0$, then the deep neural network has $L$ layers and the state of boundary neurons can be updated once every $L$ time steps.

In what follows, we will be interested in the duality mapping from the boundary space to the tangent space of trainable variables. As we shall see, the map is a composition of the activation and learning passes. 

\section{Dataset-learning duality}\label{sec:duality}

The main objective of this section is to establish a duality mapping between the tangent space of trainable variables and the boundary subspace of non-trainable variables. For starters, we consider a large neural network, defined in Sec. \ref{sec:septuple},  in a local learning equilibrium \cite{TTML}. In other words, the subject of the study will be a high-dimensional, yet local, problem that allows for significant simplifications. In particular, this allows us to reduce the effective dimensionality $K$ of the space of trainable variables to the dimensionality of the space of non-trainable variables $N$, or even to the dimensionality of the boundary subspace of non-trainable variables $N_\partial = \Tr(\hat{P})$. However, we will not consider possible symmetries in the dataset that could potentially reduce the dimensionality further.

The learning dynamics, described by Eq. \eqref{eq:learning}, can be viewed as a map from non-trainable degrees of freedom $\bf{x}$ to changes of trainable degrees of freedom $\dot{\bf{q}}$, i.e.

\be
\label{eq:learn_dyn}
({\bf f}, {\bf x}_\slashed{\partial} (t_{L-1}), ...,{\bf x}_\slashed{\partial} (t_0), {\bf x}_\partial| {\bf q}) \rightarrow \dot{\bf q}, 
\ee
where after the vertical bar there is a set of parameters of the mapping, which are fixed. So, in Eq. \eqref{eq:learn_dyn} and further vector ${\bf q}$ is the mean equilibrium value of the trainable variables vector. Similar notations containing a vertical bar cutting off fixed parameters of mapping are used further. Note that vector ${\bf f}$ makes sense of bulk neurons values after the final L-th activation step, which will be used to form the prediction of the neural network. The loss function will explicitly depend only on these values of bulk neurons. In the Eq. \eqref{eq:learn_dyn} and further notation ${\bf x}_\slashed{\partial} (t_i)$ denotes the vector of bulk neurons after $i$ steps of activation dynamics.

For example, if the loss function $H$ of trainable and non-trainable variables is separable, i.e.
\be
H = H_x({\bf x}_\partial, {\bf f}) + H_q({\bf q}),
\ee 
then evolution of weights and biases is given by
\bea
\dot{b}_{j} &=& - \gamma \delta_{j i} \frac{\partial H_q}{\partial b_i}  - \gamma  \frac{\partial {f_j}(y_j (t_{L-1}))}{\partial y_j (t_{L-1})}  \frac{\partial H_x}{ \partial {f}_j} + ..., \notag \\
\dot{w}_{j k} &=& - \gamma \delta_{j i} \frac{\partial H_q}{\partial w_{i m}} \delta_{m k} -  \gamma \delta_{k l}  x^l (t_{L-1}) \frac{\partial f_j (y_j (t_{L-1}))}{\partial y_j (t_{L-1})} \frac{\partial H_x}{ \partial {f}_j} + ...  .  \label{eq:dot_bw}
\eea
where there is no summation over $j$, $y_j (t_{L-1}) = w_{j i} x^i (t_{L-1}) + b_j$ is the total argument of the activation function $f_j$. Note that for simplicity of notation, ${\bf f}$ denotes both the argument of the loss function, formed by bulk neurons after L activation steps, and the function itself if the argument is specified in parentheses after it. So, expressions \eqref{eq:dot_bw} describe the first step in back-propagation  algorithms, ellipsis denote all other back-propagaton steps. 

The activation dynamics, in turn, makes it possible to establish a connection between the initial state of the vector of non-trainable variables $({\bf x}_\partial, {\bf x}_{\slashed{\partial}} (t_0))$ and its state at later time $({\bf x}_\partial, {\bf x}_{\slashed{\partial}} (t_i))$, i.e.
\be
\label{eq:activ_dyn}
\begin{aligned}
 {\bf x}_{\slashed{\partial}} (t_1) &= {\bf f}({\bf y}(t_0)),
\\
 {\bf x}_{\slashed{\partial}} (t_2) &= {\bf f}({\bf y}(t_1)),
\\
 ...
\\
{\bf x}_{\slashed{\partial}} (t_L) &= {\bf f}({\bf y}(t_{L-1})) = {\bf f} ({\bf x}(t_0)| {\bf {q}}).
\end{aligned}
\ee
Note that during the activation pass, the input neurons ${\bf x}_\partial$ remain fixed and act as a source for the bulk neurons ${\bf x}_{\slashed{\partial}} (t_i)$, which change with time $t_i$. As a result, the vector of bulk neurons in any activation step ${\bf x}_{\slashed{\partial}} (t_i)$ can be expressed through the vector ${\bf x}(t_0)$ before the activation starts  by recursively applying the activation function. The arguments in parentheses after ${\bf f}$ specify the functional form of the mapping, its specific form depends on variables on which it depends explicitly, they are specified as arguments. This notation is valid in the Eq. \eqref{eq:activ_dyn} and in similar cases throughout the paper.

Composition of the activation \eqref{eq:activ_dyn} and learning \eqref{eq:learn_dyn} maps is a map from non-trainable degrees of freedom $({\bf x}_\partial, {\bf x}_{\slashed{\partial}}(t_0))$ at time $t_0$ to changes of trainable degrees of freedom $\dot{\bf q}$ at time $t_L$, i.e.
\bea
({\bf x}_\partial, {\bf{x}}_{\slashed{\partial}} (t_0)| {\bf{q}}) \rightarrow \dot{\bf q}. 
\eea
For example, if the learning dynamics is described by stochastic gradient descent, then the map is given by 
\bea
\dot{q}_k({\bf x}_\partial, {\bf{x}}_{\slashed{\partial}}(t_0) |\bf{q}) & =&  - \gamma \delta_{k m} \left (\frac{\partial f_j ({\bf x}(t_0), {\bf q})}{ \partial q_m}  \frac{\partial}{\partial f_j}  + \frac{\partial}{ \partial q_m}  \right ) {H}({\bf x}_\partial, {\bf f}| {\bf q}). 
\label{eq:dot_q}
\eea
which is a map from $N$-dimensional space of non-trainable variables to $K$-dimensional space of fluctuations of trainable variables. Therefore the probability distribution $p_{\dot{q}}(\dot{\bf q})$ can be expressed as
\be
 p_{\dot{q}}(\dot{\bf q}| {\bf q}) = \int p_{x_\partial x_{\slashed{\partial}}}({\bf x}_\partial, {\bf x}_{\slashed{\partial}}(t_0))  \delta^{(K)} \left (\dot{\bf q} -   \dot{\bf q}({\bf x}_\partial, {\bf x}_{\slashed{\partial}}(t_0)| {\bf q}) \right ) \, d^{N} {\bf x}(t_0). \label{eq:dirac}
\ee  
where the different subscripts are used to emphasize that these are different probability distribution functions, e.g. $ p_{\dot{q}}()$ and $p_{x_\partial x_{\slashed{\partial}}}()$ (which is also apparent from the arguments of these functions). The vertical bar is used for conditional distribution, e.g. $ p_{\dot{q}}(\dot{\bf q}| {\bf q})$ and for emphasizing the fixed parameterization of functions, e.g.  $\dot{\bf q}({\bf x}_\partial, {\bf x}_\slashed{\partial}(t_0)| {\bf q})$. Also, $\delta^{K}$ denotes the K-dimensional Dirac delta function. Note that we also abuse notation and denote variables, $\dot{\bf q}$, and function, $\dot{\bf q}({\bf x}_\partial, {\bf x}_\slashed{\partial}(t_0)| {\bf q})$, using the same symbols. 

If the bulk neurons are initialized to zeros (or other constant values) at $t_0$ time moment before starting of activation for every dataset element, then $ p_{x_\partial x_{\slashed{\partial}}}({\bf x}_\partial, {\bf x}_{\slashed{\partial}}(t_0)) = p_{x_\partial}({\bf x}_\partial) \; \delta^{({N_\slashed{\partial}})} ({\bf x}_{\slashed{\partial}}(t_0))$, and by integrating \eqref{eq:dirac} over ${\bf x}_{\slashed{\partial}}(t_0)$ we get
\be
 p_{\dot{q}}(\dot{\bf q}| {\bf q}) = \int p_{x_\partial}({\bf x}_\partial)  \delta^{(K)} \left (\dot{\bf q} -   \dot{\bf q}( {\bf x}_\partial | {\bf x}_\slashed{\partial}(t_0), {\bf q}) \right ) \, d^{N_\partial} {\bf x}_\partial. \label{eq:dirac 1}
\ee  

If $K > N_\partial$, then we should be able to perform a local coordinate transformation
\be
\label{transf}
q^\prime_i = \Lambda^{~j}_{i} q_j.
\ee
so that $K- {N}_\partial$ direction become constraints, i.e. $\dot{q}_i^{\prime} = 0$ for $i = {N}_\partial+1, ..., K$. Note that there is more than one way to do it.

The tangent vector ${\bf{\dot{q}}}^\prime$ can be projected onto ${N}_\partial$-dimensional dynamical subspace with a projection matrix $\hat{R}$ of size $N_\partial \times K$, 
\be
\dot{q}^{\prime}_i = R^{~r}_i \Lambda^{~j}_{r} \dot{q}_j \label{eq:transformed} 
\ee
where all components are dynamical. We believe that the region, where this linear transformation of the trainable variables provides fluctuations along new $N_\partial$ coordinate axes in Euclidean space, is quite wide. Let us emphasize that although we are considering a problem local to ${\bf q}$, non-locality in the boundary space of neurons ${\bf x}_\partial$ in the general case leads us to the need to introduce curvilinear coordinate system ${\bf q}^\prime$ instead of \eqref{transf} to satisfy the requirement of fluctuations only along the $N_\partial$ coordinate axes according to considering duality.

Writing down expression similar to Eq. \eqref{eq:dirac 1} in transformed variables and integrating it over $K-{N}_\partial$ constrained directions we obtain 
\be
 p_{\dot{q}^\prime}(\dot{\bf q}^\prime| {\bf q}^\prime) = \int p_{x_\partial}({\bf x}_\partial)  \delta^{(N_\partial)} \left (\dot{\bf q}^\prime - \dot{\bf q}^\prime ({\bf x}_\partial | {\bf x}_{\slashed{\partial}} (t_0) , {\bf q}^\prime) \right ) \, d^{N_\partial} {\bf x}_\partial. \label{eq:dirac 4}
\ee 

If the map $\dot{\bf q}^\prime({\bf x}_\partial | {\bf x}_{\slashed{\partial}} (t_0) , {\bf q}^\prime)$ is invertible, then it can be considered as a true duality, and then the probability distributions are related through Jacobian matrix
\be
p_{\dot{q}^\prime}(\dot{\bf q}^\prime) = p_{x_\partial}({\bf x}_\partial({\bf \dot{q}}^\prime)) \,\left | \det \left ( \frac{\partial \dot{q}^\prime_i}{ \partial x_\partial^j} \right ) \right |^{-1}.\label{eq:prob_q3}
\ee
We shall refer to this map as the dataset-learning duality. For non-invertible maps $\dot{\bf q}^\prime({\bf x}_\partial | {\bf x}_{\slashed{\partial}} (t_0) , {\bf q}^\prime)$ we can write

\be
p_{\dot{q}^\prime}({\bf \dot{q}^\prime}) {\bf d \dot{q}^\prime} = \sum p_{x_\partial}({\bf x_{\partial}}) {\bf d x_\partial}, 
\ee
where there is summation over different ${\bf x}_\partial$ that are mapped to the same ${\bf \dot{q}}^\prime$. However, even in this more general case, the contribution from a single term in the summation might dominate (e.g., if $p_{x_\partial}({\bf x_\partial})$ dominates for some ${\bf x_\partial}$), and then Eq. \eqref{eq:prob_q3} would still be approximately satisfied.

\section{Distribution of fluctuations} \label{sec:jacobian}

To obtain an expression for the Jacobian in Eq. \eqref{eq:prob_q3}, the gradient descent equation \eqref{eq:dot_q} must be rewritten in the transformed variables \eqref{transf}, \eqref{eq:transformed}: 
\bea
\label{s_gr}
\dot{q}^\prime_i({\bf x_\partial|q}) &=& - \gamma R_i^{~r} \left( \Lambda_r^{~k} \delta_{k m} \Lambda_l^{~m} \right) \frac{d H({\bf x}_\partial, {\bf f}| {\bf q}^\prime)}{d q_m^\prime} \\
\notag &=&  - \gamma R^{~r}_i g_{r l} \left( \frac{\partial f_j({\bf x_\partial|q^\prime})}{\partial q_l^\prime} \frac{\partial H_x}{\partial f_j} + \frac{\partial H_q}{\partial q_l^\prime} \right),
\eea
where $g_{r l} = \Lambda_r^{~k} \Lambda_{l k} = \left( \Lambda \Lambda^T \right)_{rl}$. (Note that if $\Lambda$ is an orthogonal matrix, i.e. the transition to the new trainable variables is carried out through a rotation transformation, then the metric in the new variables space would remain unchanged, i.e. $g_{r l} = \delta_{r l}$.)  The Jacobian matrix can be expressed as
 \bea
\label{Jac}
 \frac{\partial \dot{q}^\prime_i}{ \partial x^k_\partial}  =  - \gamma R_i^{~r} g_{r l} \left( \frac{\partial^2 f_j({\bf x_\partial|q^\prime})}{\partial x^k_\partial ~ \partial q_l^\prime}\frac{\partial}{\partial f_j} + \frac{\partial f_j({\bf x_\partial|q^\prime})}{\partial q_l^\prime} \frac{\partial^2}{\partial x^k_\partial ~ \partial f_j} \right )  H_x({\bf x}_\partial, {\bf f}).
\eea
By substituting it back to \eqref{eq:prob_q3} we obtain an expression for the probability distribution of fluctuations of trainable variables
\be
p_{\dot{q}^\prime}(\dot{\bf q}^\prime) = \gamma^{-N} p_{x_\partial}({\bf x_\partial}({\bf \dot{q}}^\prime))  \left | \det \left( R_i^{~r} g_{r l} \left( \frac{\partial^2 f_j({\bf x_\partial|q^\prime})}{\partial x^k_\partial ~ \partial q_l^\prime} \frac{\partial}{\partial f_j} + \frac{\partial f_j({\bf x_\partial|q^\prime})}{\partial q_l^\prime} \frac{\partial^2}{\partial x^k_\partial ~ \partial f_j} \right )  H_x({\bf x}_\partial, {\bf f}) \right )\right |^{-1}  \label{eq:prob_q2}
\ee
with three factors:
\begin{enumerate}
\item $p_{x_\partial}({\bf x}_\partial)$ - distribution of non-trainable input neurons
\item $\frac{\partial H_x}{\partial f_j}$; $\frac{\partial^2 H_x}{\partial x^k_\partial \partial f_j}$  - Jacobian and Hessian  of the loss function 
\item  $\frac{\partial^2 f_j({\bf x_\partial|q^\prime})}{\partial x^k_\partial ~ \partial q_l^\prime}$; $\frac{\partial f_j({\bf x_\partial|q^\prime})}{\partial q_l^\prime}$ - dependence of the neural network (result or prediction) ${\bf f}$ on the dataset/boundary variables ${\bf x}_\partial$  and trainable  variables ${\bf q}^\prime$.
\end{enumerate}
The first factor depends directly on the boundary dynamics, i.e. training dataset, the second factor depends on the learning dynamics, i.e. the loss function, and the third factor depends on the activation dynamics, i.e. activation function. 

We recall that the transformation to primed variables \eqref{transf}, only $N_\partial$ of which fluctuate, can be carried out in different ways, i.e. the transformation matrix $\Lambda$ is not unique. In fact, we have at our disposal the entire subspace, in which the non-zero vector ${\bf \dot{q}}^\prime$ lies, to introduce an arbitrary affine coordinate system. We can carry out linear transformations in this subspace, and they will not change the form of the equations \eqref{s_gr}--\eqref{eq:prob_q2}. At the same time, we have no reason to prefer one coordinate system to another in this subspace for requirement of power-law distributions along its axes. This freedom can be used to choose the transformed (or primed) trainable variables in which the probability distribution function approximately factorizes, i.e
\be
p_{\dot{q}^\prime}(\dot{\bf q}^\prime) \approx \prod_{i=1}^{N_\partial} p_{\dot{q}_i^\prime}(\dot{ q}_i^\prime).
\ee
Then the $m$-th statistical moment for some component of the original (or unprimed) variables over some range of scales is given by
\bea
\label{moment}
\langle \dot{q}_i^m \rangle = \int \dot{q}_i^m (\dot{\bf q}^\prime) p_{\dot{\bf q}} (\dot{\bf q}(\dot{\bf q}^\prime)) \left| \frac{\partial \dot{\bf q}}{\partial \dot{\bf q}^\prime} \right| d \dot{\bf q}^\prime = \int \left( (\Lambda^{-1})_i^{~j} \dot{q}^\prime_j \right)^m p_{\dot{\bf q}^\prime}(\dot{\bf q}^\prime) d \dot{\bf q}^\prime \approx \\
\notag \approx \int \left( (\Lambda^{-1})_i^{~j} \dot{q}^\prime_j \right)^m \prod_{i=1}^{N_\partial} p_{\dot{q}_i^\prime}(\dot{ q}_i^\prime) d{q}_i^\prime. 
\eea
where the relationship between primed and unprimed variables \eqref{transf} was used.

We shall also assume that each component of fluctuation in the unprimed variables can be determined by only one single (possibly not the same for all) component in the primed variables, i.e.
\be
\dot{q}_i \approx \Lambda^{-1}_{i j} \dot{q}^\prime_j, 
\ee
where summation over $j$ is absent. Then continuing to calculate $m$-th statistical moment for $\dot{q}_i$ we get
\begin{flalign}
\label{moment_1}
\langle \dot{q}_i^m \rangle &= \int \left( (\Lambda^{-1})_{i j} \dot{q}^\prime_j \right)^m A_j \dot{q}^{\prime ~k_j}_j d \dot{q}^\prime_j = \\
& \notag = \int \left( (\Lambda^{-1})_{i j} \dot{q}^\prime_j \right)^m A_j \frac{\left((\Lambda^{-1})_{i j} \dot{q}^\prime_j \right)^{k_j}}{(\Lambda^{-1})_{i j}^{k_j + 1}} d \left((\Lambda^{-1})_{i j} \dot{q}^\prime_j \right) = \int \dot{q}_i^m p_{\dot{q}_i}(\dot{q}_i) d \dot{q}_i,
\end{flalign}
where the probability distribution of the corresponding primed variables is given by
\be
p_{\dot{q}_j^\prime}(\dot{q}_j^\prime) = A_j \dot{q}_j^{\prime ~k_j}, ~~~ \dot{q}_j^\prime \in [a, b].
\ee
Thus, a power-law distribution of fluctuations for the original trainable variable $q_i$ has the same power-law, 
\be
p_{\dot{q}_i} (\dot{q}_i) =  \frac{A_j \dot{q}_i^{k_j}}{(\Lambda^{-1})_{i j}^{k_j + 1}}, ~~~ \dot{q}_i \in \left[ (\Lambda^{-1})_{i j} a, (\Lambda^{-1})_{i j} b \right],
\ee
where the change in normalization is associated with a change in the range over which the statistical moment in unprimed variables is calculated.

\section{Toy model}\label{sec:toy}

In the previous section we considered a general multidimensional problem and applied the so-called dataset-learning duality to identify trainable variables in which the scale invariance is expected. In this section we will consider a simple example of a two-dimensional problem with both continuous and discrete degrees of freedom.

Consider a neural network consisting of only two neurons: input $x^1$ and output $x^2$ connected to with a single trainable weight $w=w_{21}$ and a bias $b=b_2$. In addition, we assume that the output neuron can take only two possible values, i.e. its marginal distribution is a sum of two delta functions, 
\be
p_{x^2}(x^2) = \frac{1}{2}\delta(x^2 - X^+) +\frac{1}{2}\delta(x^2 - X^-).
\ee
In this case, we can reduce the two-dimensional problem (i.e. two trainable variables $q_1 = w$ and $q_2 = b$ and two non-trainable variables $x^1$ and $x^2$) to two one-dimensional ones, corresponding to two different values, $x^2 = X^{+}$ and $x^2 = X^{-}$. Then for each one-dimensional problem we can define a single trainable variable $q^\prime$ that is a linear function of $q_1$ and $q_2$, and along which fluctuations will occur. Then the transformation matrices in Eqs. \eqref{transf} and \eqref{eq:transformed} are given by
\bea
\Lambda_\pm &=&
\begin{pmatrix}
\cos\theta_\pm & \sin\theta_\pm \\
-\sin\theta_\pm & \cos\theta_\pm
\end{pmatrix},\\
R &=&
\begin{pmatrix}
1 & 0
\end{pmatrix},
\eea
where the rotation angle $\theta_\pm$ corresponds to the state of the output neuron $x^2 = X^\pm$.

Let's assume that the loss function $H$ depends on  the output of the neural network $f$ after only a single step of the activation dynamics, but does not depend explicitly on $q^\prime$, i.e.
\be
H = H_x(f| x^2),
\ee
and then Eq. \eqref{s_gr} for the one-dimensional case can be written as
\be
\dot{q}^\prime  = - \gamma \frac{\partial f(x^1|{\bf q}^\prime)}{\partial q^\prime} \frac{\partial H}{\partial f}.
\ee
If we rewrite the loss function as a function of the argument $y$ of the activation function $f(y)$, i.e.
\be
y(x^1, { \bf q}^\prime) = (x^1 \cos \theta_\pm 
+ \sin \theta_\pm) q^\prime
\ee
then
\be
\label{stach_q}
\dot{q}^\prime = - \gamma \frac{\partial y}{\partial q^\prime} \frac{\partial H(y| x^2)}{\partial y} = -\gamma (x^1 \cos \theta_\pm + \sin \theta_\pm) \frac{\partial H(y|x^2)}{\partial y}.
\ee
and the expression \eqref{Jac} is given by
\be
\frac{\partial \dot{q}^\prime}{\partial x^1} = -\gamma \cos \theta_\pm \frac{\partial H(y| x^2)}{\partial y}.
\ee

Finally, we arrive at the expression for the probability distribution of fluctuations of the trainable variable in our toy model,
\be
p_{\dot{q}'}(\dot{q}') = \frac{p_{x_1}(x_1(\dot{q}'))}{\left| \gamma \cos \theta_\pm \right|} \left| \frac{\partial H(y|x_2)}{\partial y} \right|^{-1},
\ee
which is similar to the general expression \eqref{eq:prob_q2}.

\section{Emergent criticality} \label{sec:criticality}

In this section, we shall utilize the dataset-learning duality (see Sec. \ref{sec:duality}) to investigate the potential emergence of criticality arising mainly from the Jacobian matrix (see Sec. \ref{sec:jacobian}) within the context of the toy model (see Sec. \ref{sec:toy}).  The idea is to determine conditions under which the criticality might emerge in the toy model and then verify the results numerically for a toy-model classification problem. Specifically, our aim is to identify compositions of activation and loss functions that give us a power-law dependence of the Jacobian leading to a power-law distribution of fluctuations in the trainable variables. 

From conservation of probability in terms of $\dot{q}^\prime$ and $y$ we obtain
\be
p_{\dot{q}'}(\dot{q}') = p_y(y(\dot{q}')) \left| \frac{\partial y}{\partial \dot{q}'} \right|,\label{eq:criticality}
\ee
where the function $y(\dot{q}')$ is assumed to be invertible. On one hand, the power-law dependence of the Jacobian, i.e.
\be
\label{Jacobian}
\left| \frac{d y}{d \dot{q}'} \right| = \frac{1}{A |\dot{q}'|^k},~~~ A = A(w,b)>0, k\geq 0,
\ee
implies two possible differential equations
\bea
\dot{q}' = \begin{cases}
\text{sign}(\dot{q}') \exp(A y + B)  \;\;\;\;&\text{for $k =  1$}\\ 
\text{sign}(\dot{q}')  (A y + B)^{\frac{1}{1-k}}\;\;\;\;&\text{for $k \neq 1$} \label{eq:second}
\end{cases}
\eea
for some new $A = A(w,b)$, $B = B(w,b)$, and where the form of expressions for fluctuations depends on their sign. On the other hand, the gradient descent equation \eqref{stach_q} expressed through $y$ implies another differential equation

\be
\label{grad}
\dot{q}' = - (D y - C) \frac{\partial H}{\partial q^\prime},
\ee
where
\be
\begin{aligned}
    C &= \frac{\gamma \cos (\theta_\pm)}{w} (b - w \tan (\theta_\pm)), \\
    D &= \frac{\gamma \cos (\theta_\pm)}{w}.
\end{aligned}
\label{CD}
\ee
and for the gradient descent 
\be 
D y - C > 0 \label{eq:sign}.
\ee
In this section we shall study different compositions of loss and activation functions, i.e. $H(f(y))$, for which equations \eqref{eq:second} and \eqref{grad} are satisfied and thus the emergence of criticality is expected. 

For $k = 1$ equations \eqref{eq:second} and \eqref{grad} can be combined together as
\be
\frac{d H(f(y))}{d y} = - \text{sign}(\dot{q}') \frac{\exp (A y + B)}{D y - C},
\label{eq:exponential}
\ee
By changing variable $z = A y - A C/D$ in  \eqref{eq:exponential} and integrating in some region with respect to $z$ we obtain 
\be
H(f(z)) = -\text{sign}(\dot{q}')  \frac{1}{D} \exp \left ( \frac{AC}{D} + B \right ) \int_{z_0}^{z} dz'  \frac{\exp (z')}{z'},
\label{eq:exponential}
\ee
or 
\be
H(f(y)) = -\text{sign}(\dot{q}') \frac{1}{D} \exp \left ( \frac{AC}{D} + B \right ) \int^{A y - A C/D}_{z_0} dz'  \frac{\exp (z')}{z'},
\label{eq:exponential}
\ee 
One can show (see Appendix \ref{sec:integral}) that for $|z| \gg 1$ the integral $\int d z^\prime \frac{\exp (z^\prime)}{z^\prime}$ can be approximated by the integrand, i.e.
\be
H(f(y)) \approx -\text{sign}(\dot{q}') \frac{\exp(A y + B)}{A (D y - C)}.
\ee
An arbitrary constant coming from the lower limit of the integral must also be added to the expression. We omit it here and in the following cases for simplicity.

Note that in the limit $|D y | \ll |C|$, $|A| \gg |D/C|$ and $B = \log |A C|$ we get an exponential function 
\be
\boxed{H(f(y)) = \exp (A y)}.
\label{exp_comp}
\ee

For $k \in [0,1)$ equations \eqref{eq:second} and \eqref{grad} can be combined together as
\bea
\frac{d H(f(y))}{d y} &=& - \text{sign}(\dot{q}') \frac{(A y +B)^\alpha}{D y-C} \\
&=& - \text{sign}(\dot{q}') \frac{A}{D} \frac{(A y+ B)^\alpha}{(A y + B) - \left(B + \frac{A C}{D}\right)},\notag
\eea
where $\alpha \equiv 1/(1-k) > 1$. If
$|A y +B| \gg \left|B + \frac{A C}{D} \right|$, then
\be
\frac{d H(f(y))}{d y} = - \text{sign}(\dot{q}') \frac{A}{D} \frac{(A y+ B)^\alpha}{A y + B} = - \text{sign}(\dot{q}')  \frac{A}{D} (A y + B)^{\alpha - 1},
\ee
whose solution is given by
\be
H(f(y)) =- \text{sign}(\dot{q}') \frac{A^\alpha}{\alpha D} \left(y + \frac{B}{A} \right)^\alpha. 
\ee
Note that for $\left|\frac{A^\alpha}{\alpha D}\right| =1$, $\Delta = \frac{B}{A}$ we get the following power law-dependence
\be
\boxed{H(f(y)) = (y + \Delta)^\alpha}.
\label{power_comp1}
\ee
In the opposite limit, i.e. $|A y +B| \ll \left|B + \frac{A C}{D} \right|$, we get
\be
\frac{d H(f(y))}{d y} = \text{sign}(\dot{q}')\frac{A}{D} \frac{(A y+ B)^\alpha}{B + \frac{A C}{D}},
\ee
or 
\be
H(f(y)) = \text{sign}(\dot{q}')\frac{A^{\alpha + 1}\left(y + \frac{B}{A}\right)^{\alpha +1}}{(\alpha +1)(A C + B D)}.
\ee
For $\left| \frac{A^{\alpha + 1}}{(\alpha +1)(A C + B D)} \right|= 1$ and $\Delta = \frac{B}{A}$ the desired composition function takes the form
\be
\boxed{H(f(y)) = (y + \Delta)^{\alpha + 1}}.
\label{power_comp2}
\ee

For $k=2$ or $\alpha =1/(1-k) = -1$ equations \eqref{eq:second} and \eqref{grad} can be combined together as
\be
\frac{d H(f(y))}{d y} =  \frac{\text{sign}(\dot{q}') }{(A y + B)(C - D y)} = \frac{\text{sign}(\dot{q}') }{-A D y^2 + (A C - B D) y + B C},
\ee
If the quadratic term in the denominator is small, i.e. $|A D y| \ll |A C - B D|$, then upon integration we obtain
\be
H(f(y)) = \text{sign}(\dot{q}') \frac{ \log |(AC - BD) y + BC|}{A C - B D}.
\ee
For $|A C - B D| = 1$, $\Delta = B C$ we get the following form of composition function
\be
\boxed{H(f(y)) = - \log |\Delta - y|}.
\label{log_comp}
\ee

\section{Numerical results}\label{sec:numeric}

In this section, we shall present numerical results for the toy model described in Sec. \ref{sec:toy}, and compare them with analytical results obtained in Sec. \ref{sec:criticality}. In particular, we are interested in studying the emergence of critical behavior, or when the distribution of changes in trainable rotated variable $q^\prime$ is described by a power-law, i.e.,
\be
p(\dot{q}^\prime) \propto (\dot{q}^\prime)^{-k}\label{eq:criticality1}.
\ee
The goal is to provide specific numerical examples of the critical behavior \eqref{eq:criticality} on some ranges of scales for exponential \eqref{exp_comp}, power-law \eqref{power_comp2} and logarithmic \eqref{log_comp} compositions of activation and loss functions, i.e. for $H(f(y))$.

For the numerical experiments we consider a simple dataset with only two classes (called `$-$' and `$+$') with a single output neuron which takes discrete values ($x^2= X^- = 0$ or $x^2= X^+ = 1$). The input subspace also consists of only a single neuron that takes continuous values drawn from two Gaussian probability distributions $p_-(x^1)$ (for $x^2=0$) and $p_+(x^1)$ (for $x^2=1$) with mean 
\bea
\langle x^1 \rangle_- &=& 0 \label{eq:mean}\\
\langle x^1 \rangle_+ &=& 1\notag
\eea
where notations $\langle x^1 \rangle_-$ and $\langle x^1 \rangle_+$ mean averaging over the distributions of the values of the input neuron $x^1$ with conditions $x^2 = X^-$ and $x^2 = X^+$, respectively. The standard deviations are chosen as follows 
\bea\label{eq:std}
\sqrt{\langle (x^1)^2 \rangle_- - \langle x^1 \rangle_-^2}  &=& \sqrt{\langle (x^1)^2 \rangle_- } = 0.25 \\
\sqrt{\langle (x^1)^2 \rangle_+ - \langle (x^1) \rangle_+^2} &=& \sqrt{\langle (x^1)^2 \rangle_+ - 1}  =  0.25.\notag
\eea
This simple dataset allows us to use the architecture of the toy model of Sec. \ref{sec:toy} with one input neuron $x^1$, one output neuron $x^2$, one trainable bias $b$ and one trainable weight $w$ (from input neuron to output neuron). 

Note that the main goal of this analysis is not to obtain high prediction accuracy (although it is above $97\%$ in all experiments), or to develop an architecture for a realistic classification problem (although a similar dataset would have been obtained if we considered a truncated MNIST dataset \cite{MNIST} with only images of `zeros' and `ones' and input states projected down to a single dimension). Rather, our aim is to demonstrate the emergence of criticality within a simple low-dimensional problem, but we expect the same mechanisms to be responsible for the emergence of criticality in higher-dimensional problems \cite{SOC}. 

Also, note that in the learning equilibrium, the average value of the trainable parameters does not change significantly. Therefore, the distribution of the actual changes of trainable variables, i.e., $p(\dot{q})$, and the distribution of potential jumps when the trainable parameters remain fixed, i.e., $p\left (- \gamma \frac{\partial H}{\partial q}\right )$, are essentially the same. However, the latter is more suitable, i.e. we will consider potential jumps caused separately by `$-$' and `$+$' classes, which, in the learning equilibrium, can be described as a two one-dimensional problem. 

In the remainder of the section we shall consider five experiments, two for exponential \eqref{exp_comp}, two for power-law \eqref{power_comp2}, and one for logarithmic \eqref{log_comp} composition of activation and loss functions.

\subsection{Exponential composition, $k=1$}

Consider sigmoid activation function
\be
\label{sigm}
f = (1 + e^{-y})^{-1}
\ee
with mean-squared loss
\be
\label{srcv}
H(f, x^2) = (f - x^2)^2
\ee
or cross-entropy loss 
\be
\label{cren}
H(f, x^2) = -(1-x^2) \log(1-f) -x^2 \log(f).
\ee

\begin{figure}[H]
    \centering  
    \begin{subfigure}[t]{0.49\textwidth} 
        \centering
        \includegraphics[width=\linewidth]{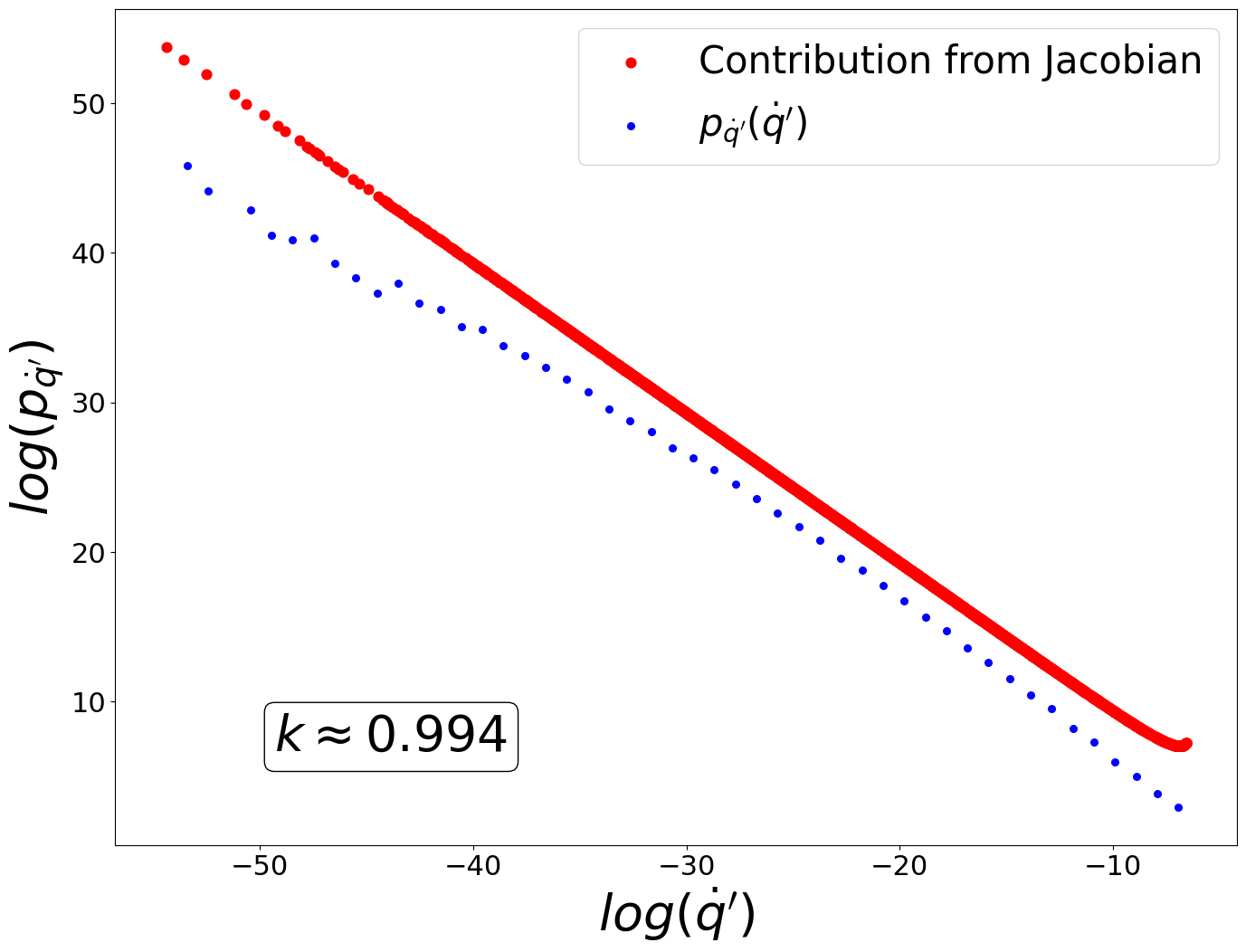}
        \caption{`$-$' class}
    \end{subfigure}
    \hfill
    \begin{subfigure}[t]{0.49\textwidth} 
        \centering
        \includegraphics[width=\linewidth]{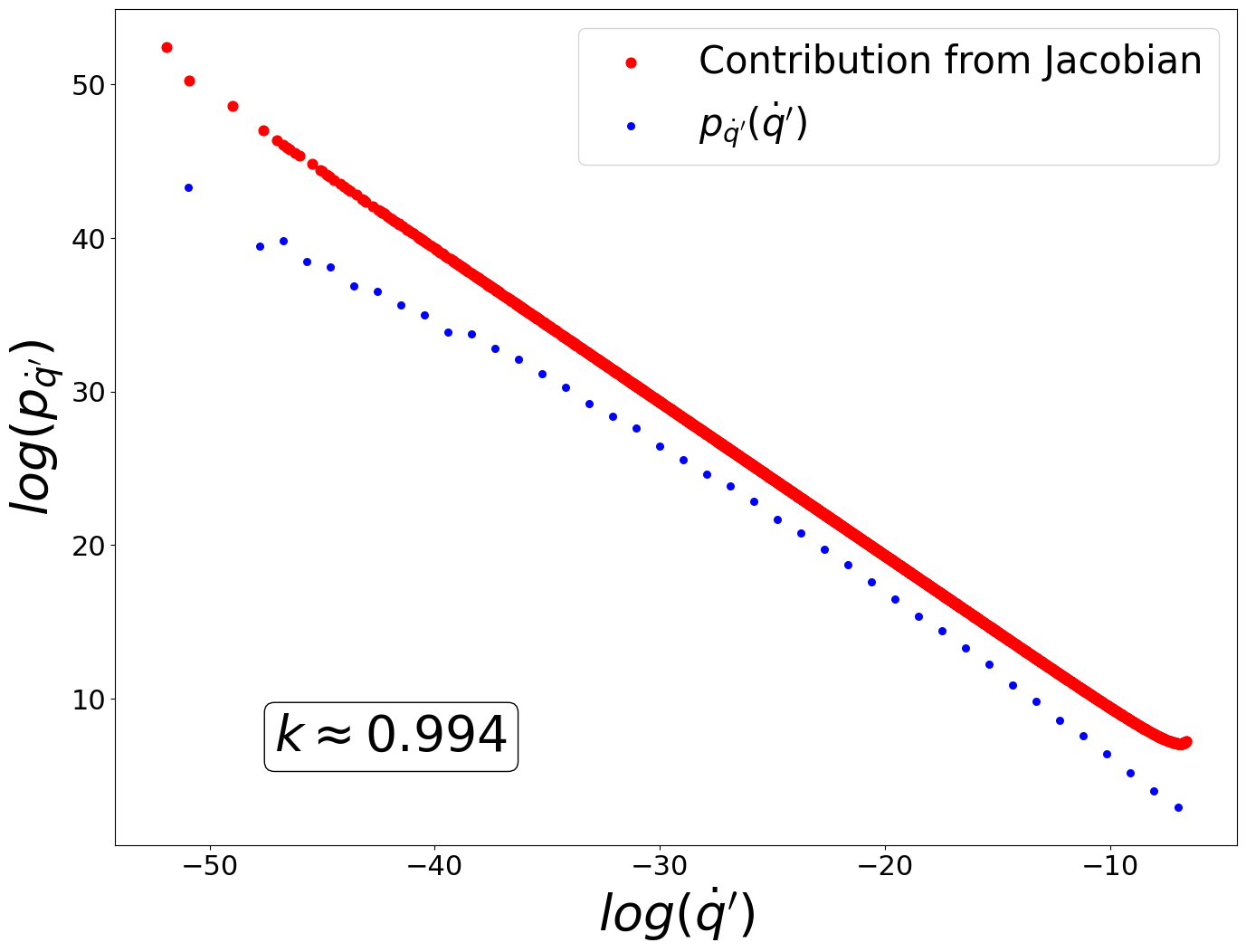}
        \caption{`$+$' class}
    \end{subfigure}
    \caption{Distribution of  fluctuations for composition of sigmoid activation and mean squared loss functions.}
    \label{fig:graphs_mean_squared}
\end{figure}

For composition of sigmoid activation \eqref{sigm} and mean-squared loss \eqref{srcv} functions and `$-$' class, in the limit $y < 0$ Eq. \eqref{sigm} reduces to $f \approx e^{y}$
and then the composition function is
    \be \label{loss_ms_0}
    H(f(y)) \approx e^{2 y}.
    \ee
Likewise, for the same composition and `$+$' class, in the limit $y > 0$ Eq. \eqref{sigm} reduces to $f \approx 1 - e^{-y}$
and then composition function is
    \be \label{loss_ms_1}
    H(f(y)) \approx e^{-2 y}.
    \ee
In Fig. \ref{fig:graphs_mean_squared}, we plot in logarithmic axes the distributions of fluctuations of $q'$ (blue dots) and the contribution from the Jacobian (red dots) to these distributions corresponding to Eq. \eqref{eq:criticality}. The linear behavior with slope $-1$ (or $k=1$) is in good agreement with Eq. \eqref{exp_comp}.

\begin{figure}[H]
    \centering  
    \begin{subfigure}[t]{0.49\textwidth} 
        \centering
\includegraphics[width=\linewidth]{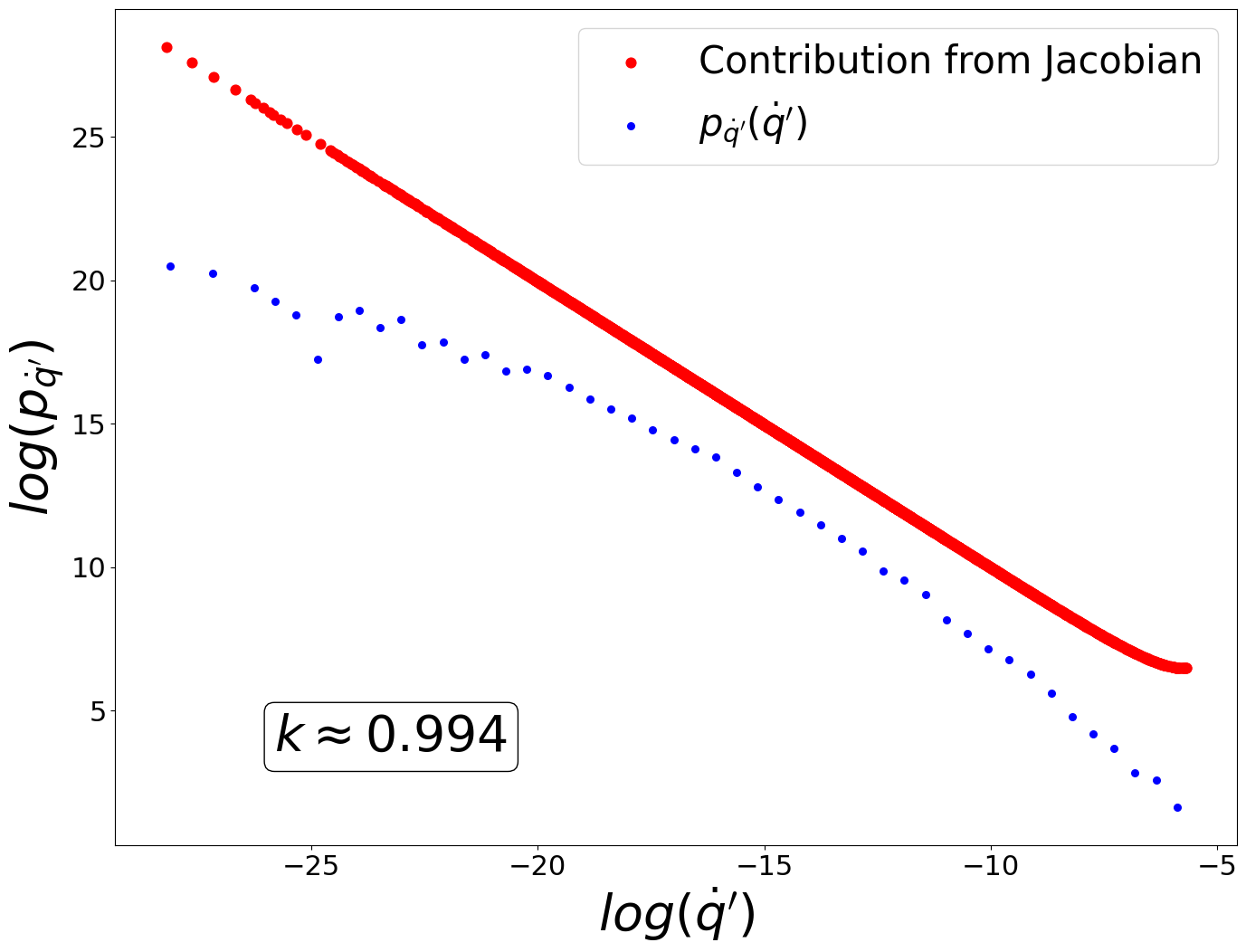}
        \caption{`$-$' class}
    \end{subfigure}
    \hfill
    \begin{subfigure}[t]{0.49\textwidth}
        \centering
 \includegraphics[width=\linewidth]{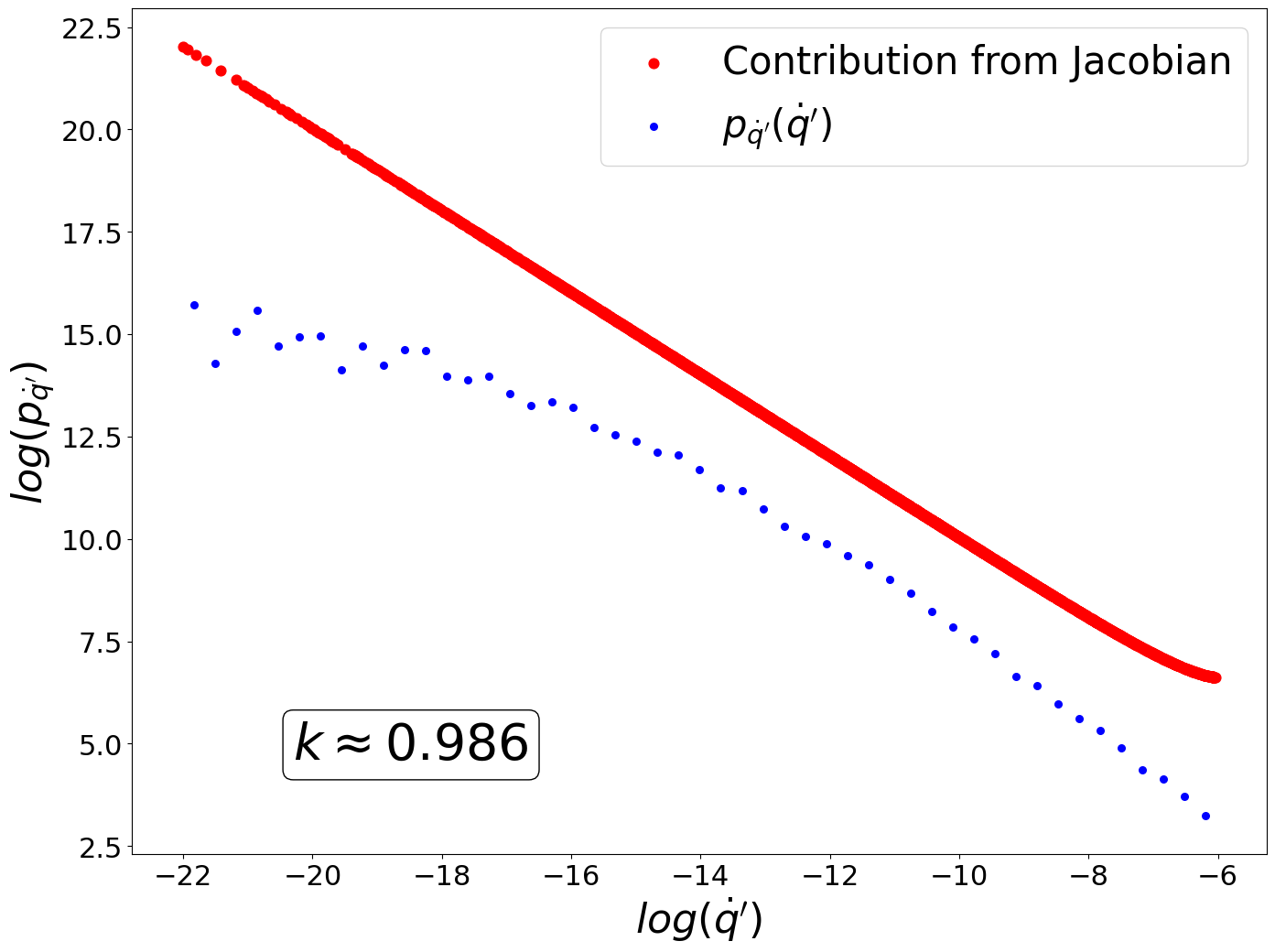}
        \caption{`$+$' class}
    \end{subfigure}
    \caption{Distribution of  fluctuations for composition of sigmoid activation and cross entropy loss functions.}
    \label{fig:graphs_sigmoid_cross}
\end{figure}

For composition of sigmoid activation \eqref{sigm} and cross-entropy loss \eqref{cren} function we obtain, for `$-$' class and in the limit $y < 0$, 
   \be
   H(f(y)) = - \log(1-f(y)) \approx e^y, 
   \ee
and for `$+$' class and in the limit  $y > 0$,
   \be
H(f(y)) = \log(f(y)) \approx e^{-y}. 
   \ee
In Fig. \ref{fig:graphs_sigmoid_cross}, we plot the distribution of fluctuations of $q'$ which is once again in agreement with Eq. \eqref{exp_comp}. Note that in this case, despite a nearly perfect power-law contribution from the Jacobian (red dots), the distribution of fluctuations as a whole (blue dots) is significantly distorted by the contribution from $p_y(y(\dot{q}^\prime))$ in Eq. \eqref{eq:criticality}.
 
We conclude that for the toy-model classification problem with a sigmoid activation function and either mean squared error or cross-entropy loss functions, the fluctuations of trainable variables can follow a power law with $k=1$, as confirmed both numerically and analytically.

\subsection{Power-law composition, $k=0; \frac{2}{3}$}

Consider a composition of a ReLU activation function 
\be \label{relu_ms}
    f (y) = \max(0, y)
\ee
and the power-law loss function
\be
    H(f,x^2) = (f - x^2)^n \label{eq:msl}.
\ee
In this case, non-vanishing fluctuations can take place only when the argument of ReLU is positive, i.e.
\be
y > 0,
\ee
then the composition function is also power-law 
\be
H(f(y)) = (y - x^2)^n.\label{eq:power-loss}
\ee

\begin{figure}[H]
    \centering  
    \begin{subfigure}[t]{0.49\textwidth} 
        \centering
        \includegraphics[width=\linewidth]{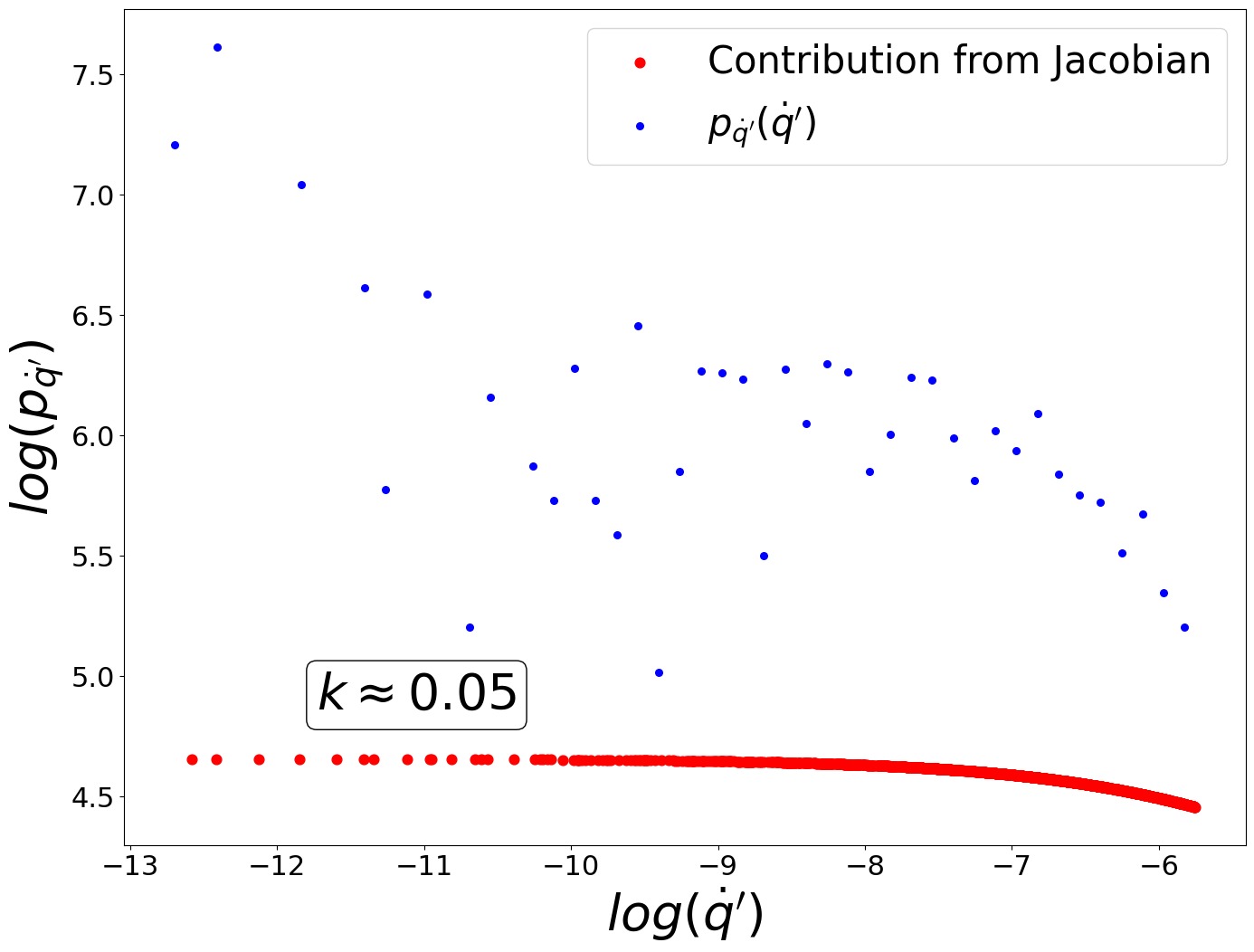}
        \caption{`$-$' class}
    \end{subfigure}
    \hfill
    \begin{subfigure}[t]{0.49\textwidth} 
        \centering
        \includegraphics[width=\linewidth]{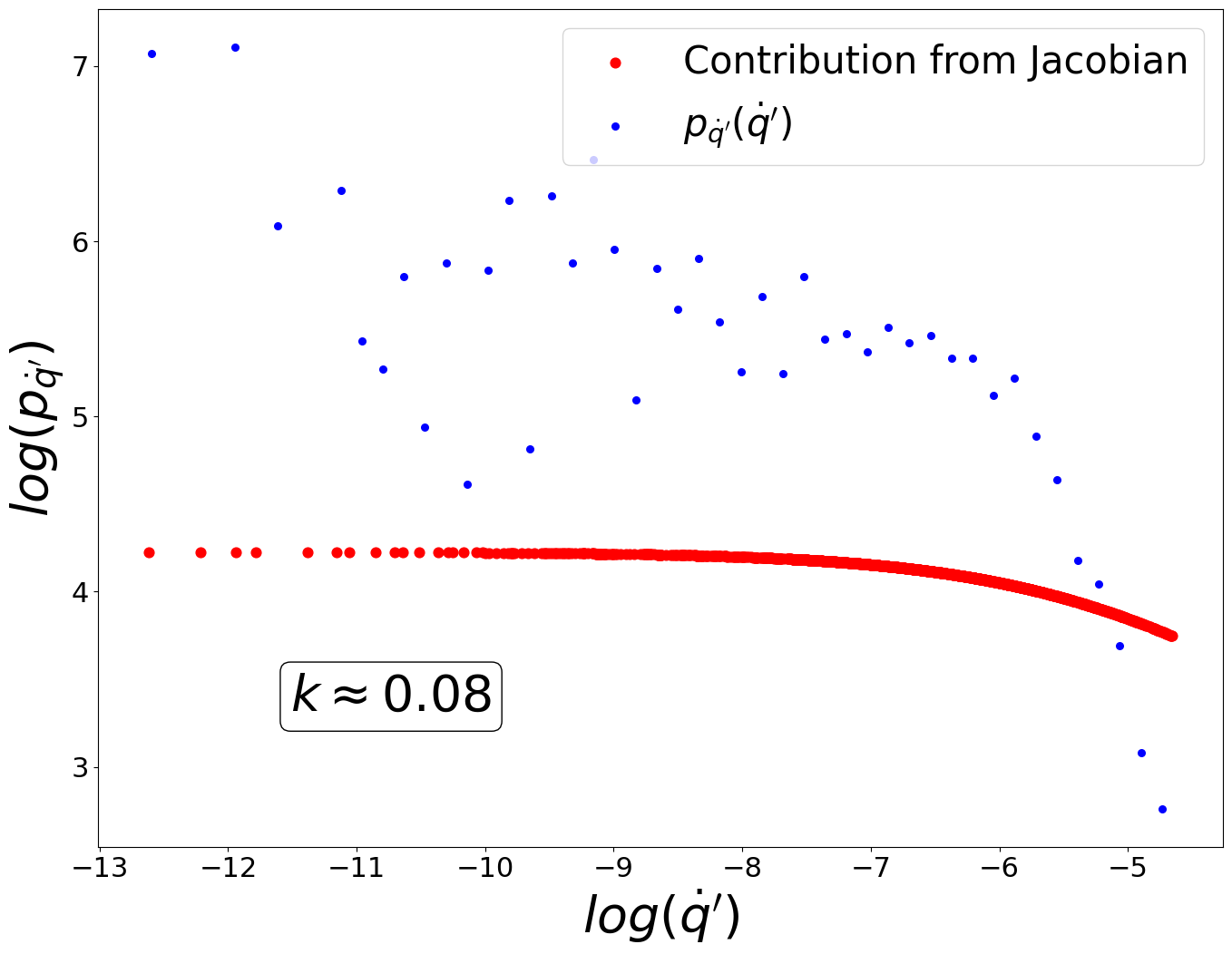}
        \caption{`$+$' class}
    \end{subfigure}  
    \caption{Distribution of  fluctuations for composition of sigmoid activation and cross entropy loss functions.}
    \label{fig:graphs_relu_ms}
\end{figure}

For $n = 2$, i.e., the standard mean squared loss, Eq. \eqref{power_comp2} implies that 
\be
k = \frac{\alpha - 1}{\alpha} = \frac{n-2}{n-1} = 0,
\ee
which is in agreement with the numerical results plotted in Fig. \ref{fig:graphs_relu_ms}.

\begin{figure}[H]
    \centering  
    \begin{subfigure}[t]{0.49\textwidth} 
        \centering
        \includegraphics[width=\linewidth]{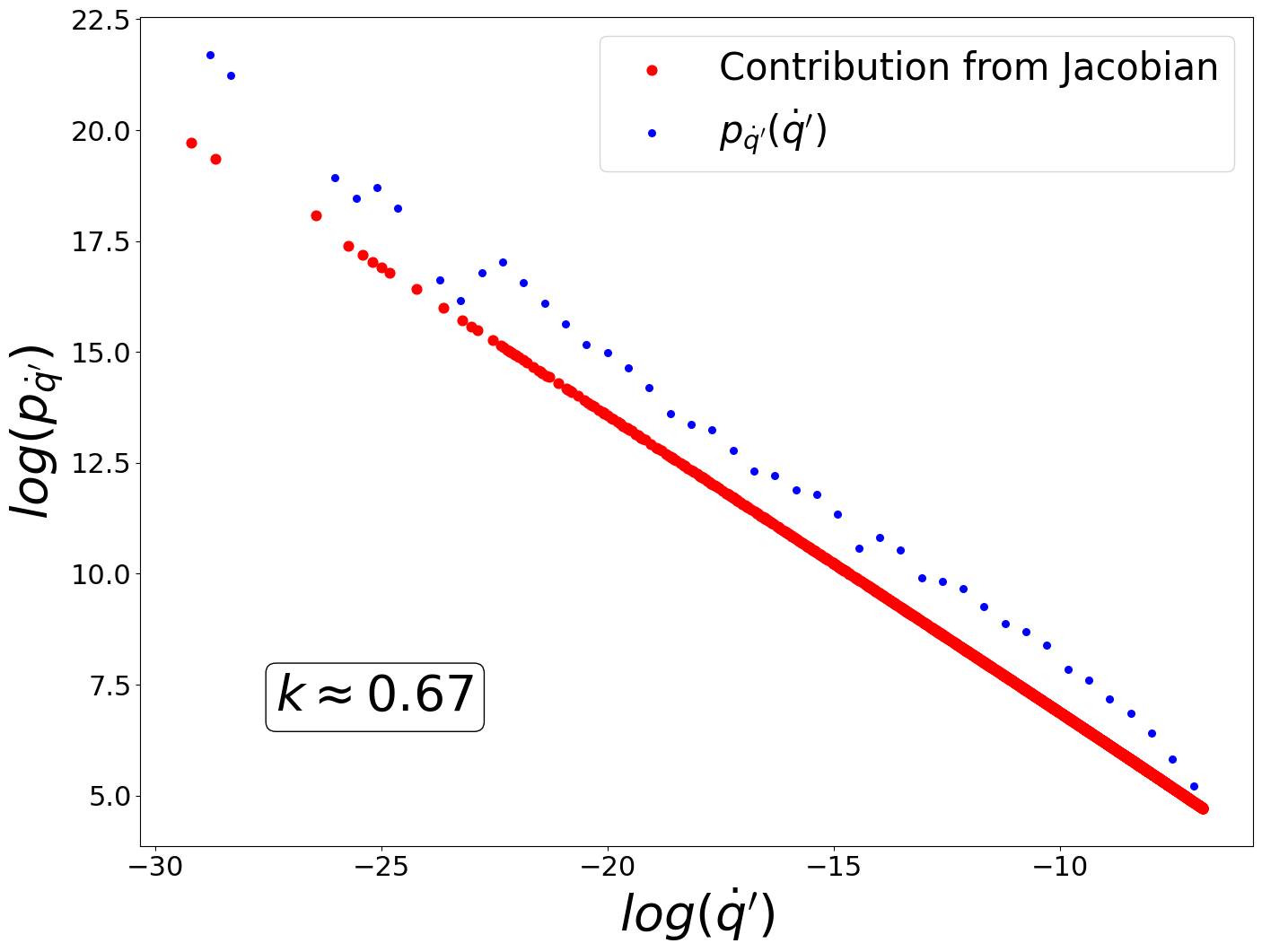}
        \caption{`$-$' class}
    \end{subfigure}
    \hfill
    \begin{subfigure}[t]{0.49\textwidth} 
        \centering
        \includegraphics[width=\linewidth]{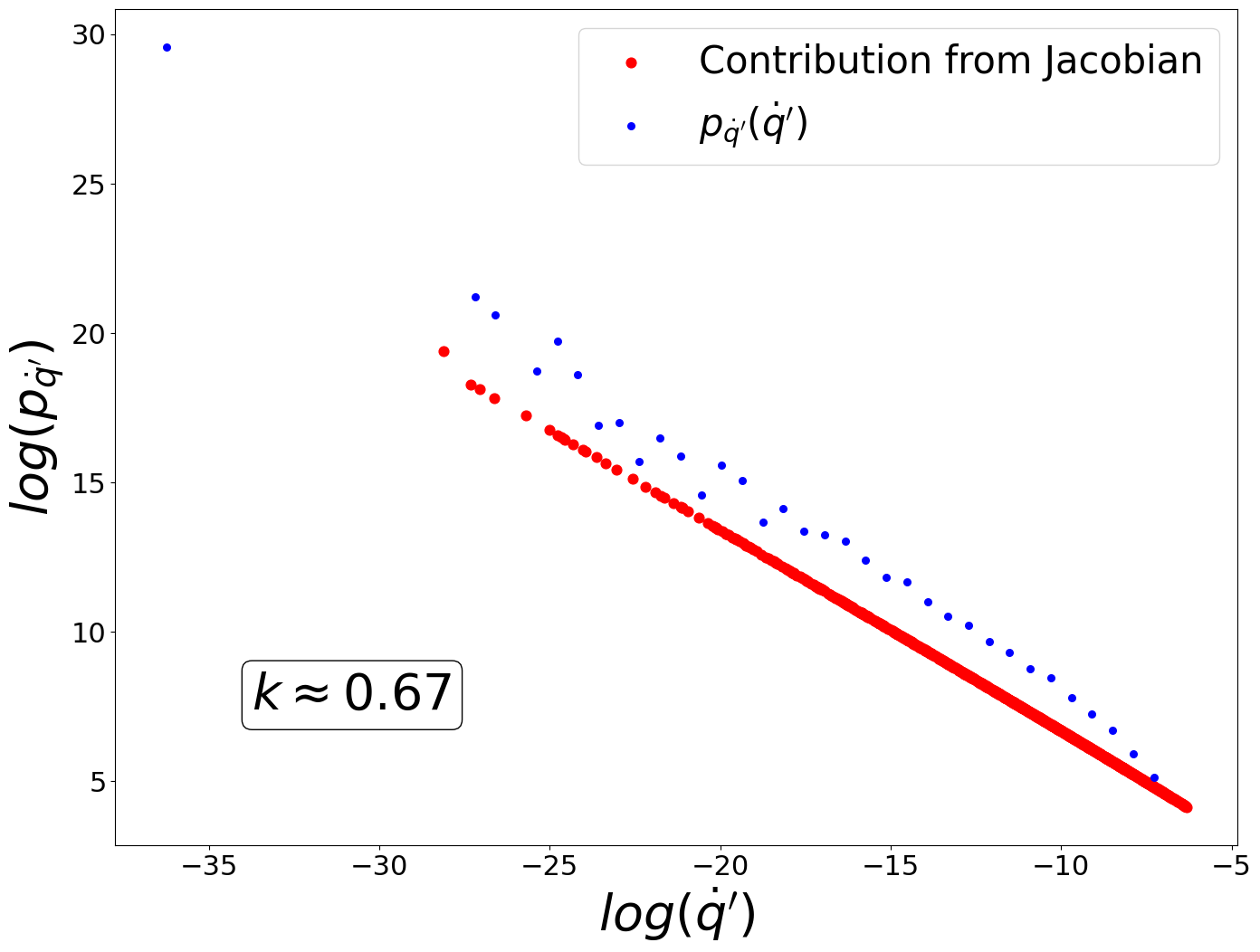}
        \caption{`$+$' class}
    \end{subfigure}
    \caption{Distribution of  fluctuations for composition of sigmoid activation and cross entropy loss functions.}
    \label{fig:graphs_relu_four}
\end{figure}

For $n = 4$, Eq. \eqref{power_comp2} implies that $k = \frac{2}{3}$ in agreement with the numerical results plotted in Fig. \ref{fig:graphs_relu_four}.

Thus, we have demonstrated that the power-law composition case \eqref{power_comp2} is achieved by combining power-law loss and ReLU activation functions. More generally, we can obtain any $k = \frac{n-2}{n-1}$, where $n$ is the power that appears in \eqref{eq:power-loss}.

\subsection{Logarithmic composition, $k=2$} 

Consider the cross-entropy loss function \eqref{cren} and a piece-wise linear activation function
\be \label{piecewice}
f(y) = \begin{cases}
        y, ~0 < y < 1,
        \\
        0,~ y < 0 \text{~or~} y > 1.
    \end{cases}
\ee
Within the range $0 < y <1$, the composition of activation and loss function is
\be
H(f(y)) = -(1 - x_2) \log(1-y) - x_2 \log(y).
\ee
which is the logarithm composition of Eq. \eqref{log_comp} for `$-$' class (i.e. $x^2 = 0$) with $|\Delta| = 1$, and for `$+$' class (i.e. $x^2 = 1$) with $\Delta = 0$.

\begin{figure}[H]
    \centering  
    \begin{subfigure}[t]{0.49\textwidth} 
        \centering
        \includegraphics[width=\linewidth]{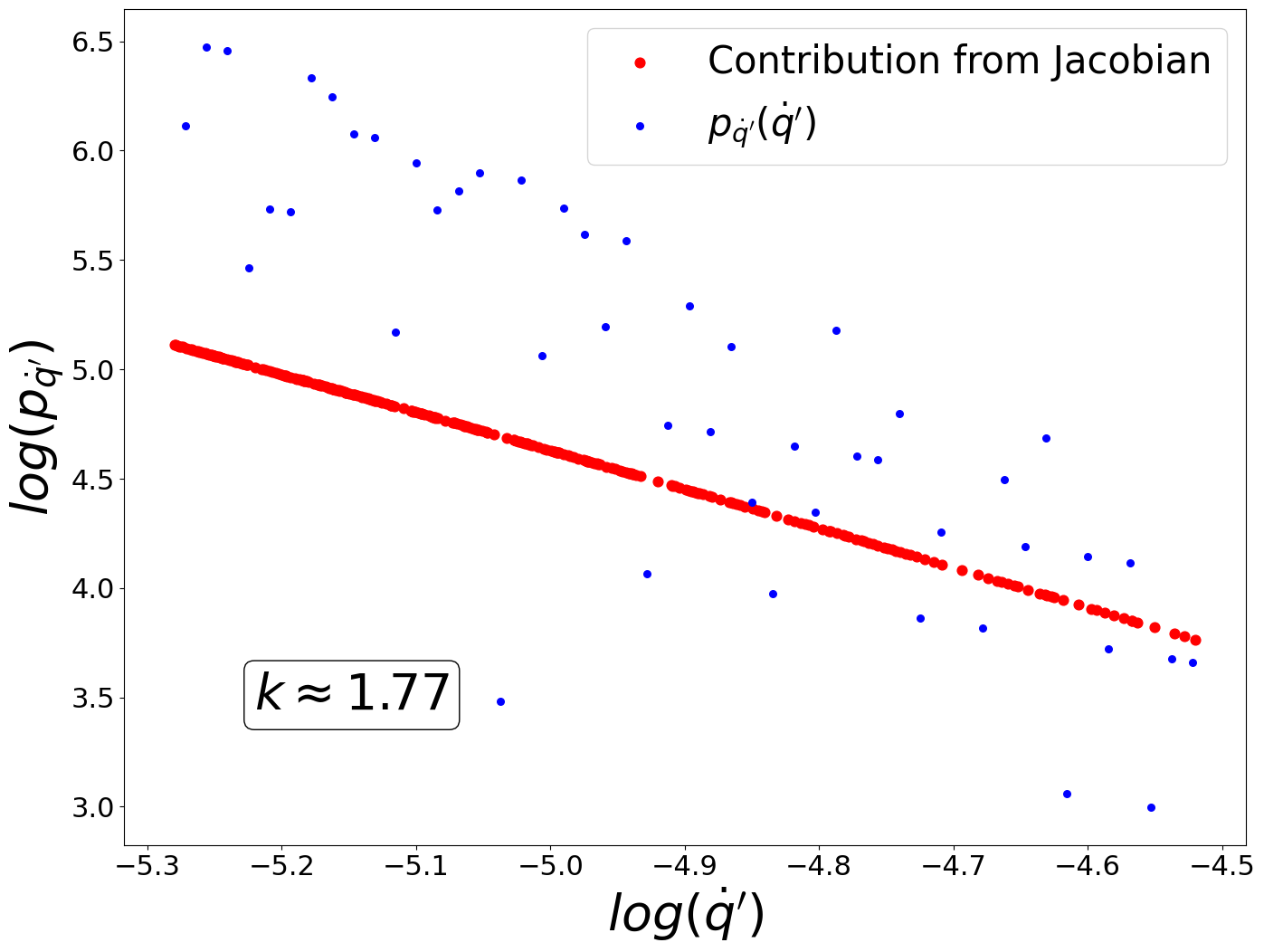}
        \caption{`$-$' class}
    \end{subfigure}
    \hfill
    \begin{subfigure}[t]{0.49\textwidth} 
        \centering
        \includegraphics[width=\linewidth]{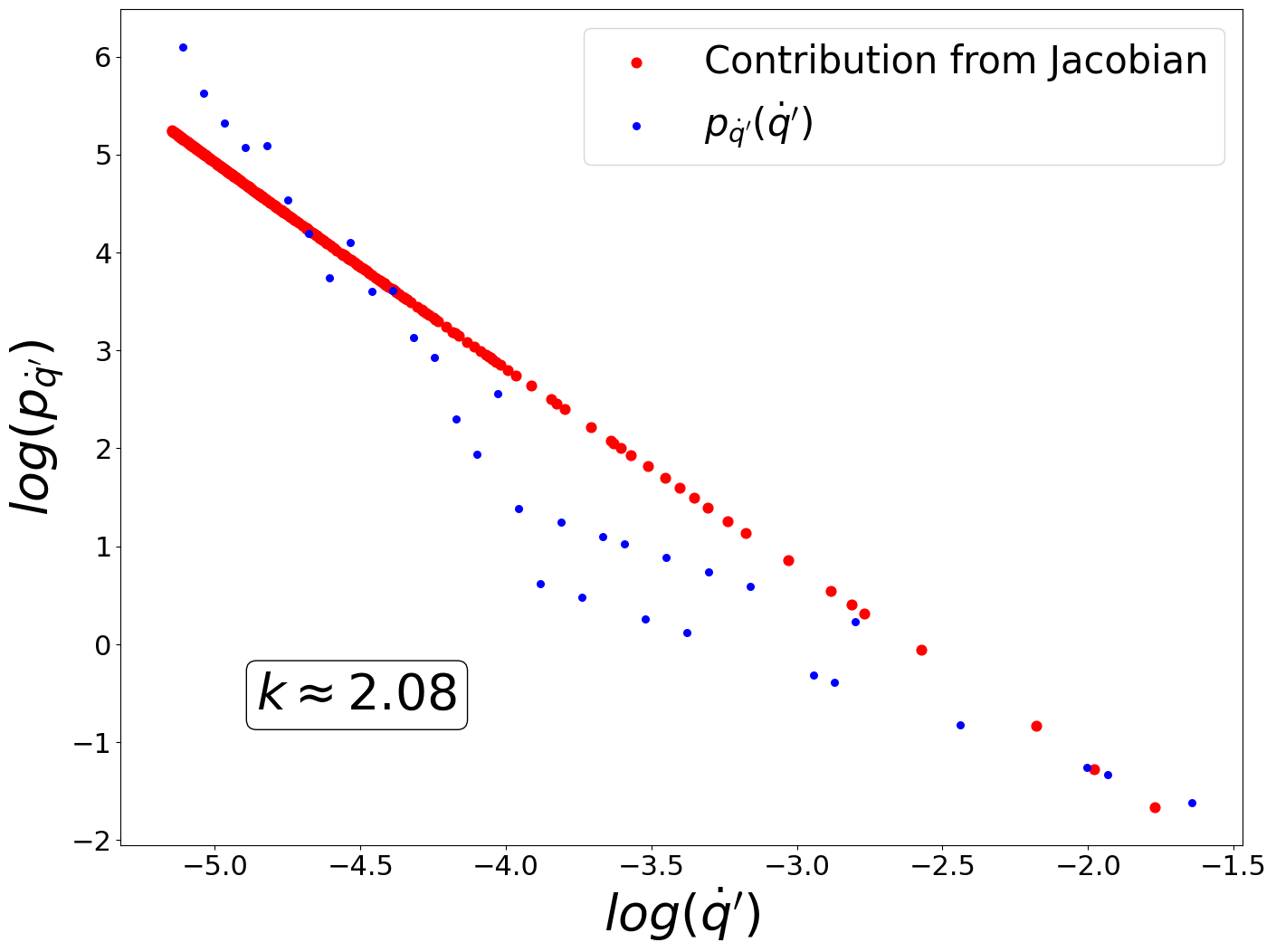}
        \caption{`$+$' class}
    \end{subfigure}
    \caption{Distribution of  fluctuations for composition of sigmoid activation and cross entropy loss functions.}
    \label{fig:graphs_piece_log}
\end{figure}

In Fig. \ref{fig:graphs_piece_log}, we plot the distribution of fluctuations of $q'$, for which $k=2$, in good agreement with the analytical results in \eqref{log_comp}.

\section{Discussion}\label{sec:Results}

In this paper, we accomplished two main tasks. 

Firstly, we established a duality mapping between the space of boundary neurons (i.e., the dataset) and the tangent space of trainable variables (i.e., the learning), the so-called dataset-learning duality. Both spaces (the boundary and the tangent) are generally very high-dimensional, making the analysis of the duality very non-trivial. However, by considering the problem in a local learning equilibrium and under the assumption that the probability distribution function of the tangent space variables is factorizable, the multidimensional problem can be greatly simplified. 

Secondly, we applied the dataset-learning duality to study the emergence of criticality in the learning systems. We show that the observed scale-invariance of fluctuations of the trainable variables (e.g., weights and biases) is caused by the emergence of criticality in the dual tangent space of trainable variables. In particular, by considering decoupled one-dimensional problems we analyzed different compositions of activation and loss function which can give rise to a power-law distribution of fluctuations of trainable variables on a wide range of scales. We showed that the power-law, exponential and logarithmic compositions of activation and loss functions can all give rise to criticality in the learning dynamics even if the dataset is in a non-critical state. Main results of the study of criticality are summarized the following table.

\begin{center}
\begin{tabular}{ | m{3 cm} |m{4cm}| m{6cm} | } 
  \hline
  $k$& $H(f(y))$& Examples \\ 
  \hline
    1 & $\exp (A y) $ & Cross-entropy or mean squared loss functions and sigmoid activation function \\ 
  \hline
  $\frac{\alpha -1}{\alpha}$ & $(y + \Delta)^{\alpha + 1}$ & Power-law loss and ReLU activation functions\\ 
  \hline
   $2$ & $-\log |\Delta - y|$ & Cross-entropy loss and piece-wise linear activation functions \\ 
  \hline
\end{tabular}
\end{center}

Besides its theoretical significance and potential relevance in modeling critical phenomena in physical and biological systems \cite{Landau, Romanenko}, the emergence of criticality is expected to play a central role in machine learning applications. Power-law distributions, in particular, enable trainable variables to explore a broader range of scales without facing exponential suppression.  Consequently, criticality is presumed to prevent neural networks from becoming trapped in local minima, a highly desirable property for any learning system. However, the analysis of the learning efficiency and its relation to criticality must involve considerations of non-equilibrium systems, which is beyond the scope of the current paper. Nevertheless, we expect that the established dataset-learning duality can be developed further to shed light on non-equilibrium problems. We leave these and other related questions for future research.

{\it Acknowledgements.} The authors are grateful to Yaroslav Gusev for his invaluable assistance with both numerical and analytical problems. His expertise was crucial to the success of this project. 

\bibliographystyle{unsrt}
\bibliography{library.bib}

\appendix

\section{Approximation of an integral} \label{sec:integral}

Consider the following integral 
\be
I(x) = \int^x \frac{\exp (z)}{z} d z.
\ee

In the limit of $x \gg 1$, we can integrate by by parts (iteratively) to obtain 
\be
I(x) = \frac{\exp (x)}{x} + \frac{\exp (x)}{x} \sum_{n =1}^\infty \frac{n!}{x^n} \approx \frac{\exp (x)}{x} + \frac{\exp (x+1)}{x^2} \approx \frac{\exp (x)}{x},\label{eq:a1}
\ee
where we took into account that $\sum_{n =1}^\infty \frac{n!}{x^n} \approx \frac{e}{x \left( 1 - \frac{1}{x} \right)}$.

In an opposite limit $y \equiv -x \gg 1$ we get
\be
I(y) = - \frac{\exp (-y)}{y} + \frac{\exp (-y)}{y} \sum_{n = 1}^\infty \frac{(-1)^{n+1}n!}{y^n} \approx - \frac{\exp (-y)}{y} + \frac{\exp (-y-1)}{y^2} \approx - \frac{\exp (-y)}{y},\label{eq:a2}
\ee
where we took into account that $\sum_{n = 1}^\infty \frac{(-1)^{n+1}n!}{y^n} \approx \frac{e^{-1}}{y \left( 1 + \frac{1}{y} \right)}$.

By combining \eqref{eq:a1} and \eqref{eq:a2} we obtain
\be
I(x) \approx \frac{\exp (x)}{x}.
\ee
for $|x| \gg 1$.

\end{document}